\pdfoutput=1
\documentclass[onecolumn]{fairmeta}
\usepackage{amsmath}
\usepackage{amssymb}
\usepackage[utf8]{inputenc} % allow utf-8 input
\usepackage[T1]{fontenc}    % use 8-bit T1 fonts
\usepackage{hyperref}       % hyperlinks
\usepackage{url}            % simple URL typesetting
\usepackage{booktabs}       % professional-quality tables
\usepackage{amsfonts}       % blackboard math symbols
\usepackage{nicefrac}       % compact symbols for 1/2, etc.
\usepackage{microtype}      % microtypography
\usepackage{xcolor}         % colors
\usepackage{graphicx}
\usepackage{multirow}
\usepackage{makecell}
\usepackage{colortbl}
\definecolor{morandipink}{HTML}{FFE1EA}
\definecolor{morandiyellow}{HTML}{FFF2CC}

\title{ActiveVital: Geometry-Aware Embodied Vital Signs Monitoring for Home Healthcare Robots}

\author[1,2]{Yuxuan Hu}
\author[1]{Shihao Li}
\author[1]{Yang Xiao}
\author[1]{Gen Li}
\author[2]{Feng Xu}
\author[1]{Jianfei Yang}

\affiliation[1]{MARS Lab, Nanyang Technological University, Singapore}
\affiliation[2]{Fudan University, Shanghai, China}

\abstract{
Home robots require reliable vital signs monitoring to support long-term companionship and safety in daily environments, yet obtaining respiration and heart rate without physical contact remains challenging in unconstrained home settings.
Millimeter-wave (mmWave) radar offers a promising solution due to its phase sensitivity to sub-millimeter motions. However, mmWave measurements are fundamentally constrained by observation geometry, since only the radial component of motion is observable. Consequently, arbitrary robot--human orientations often introduce angular misalignment that destabilizes vital signs estimation.
To address this limitation, we reformulate vital signs monitoring from passive signal recovery to active geometric regulation. We propose ActiveVital, a vision-guided sensing framework that treats sensing geometry as an explicit control variable for robots.
It localizes the chest anchor via visual keypoints and converts alignment errors into control commands.
This steers the robot-mounted radar toward near-normal incidence to the thoracic surface, maximizing radial observability within a perception-action loop.
A differential phase enhancement module further stabilizes signal extraction under motion.
Experiments show that ActiveVital reduces respiration interval error from 0.87\,s to 0.14\,s and heart rate error from 13.59\,bpm to 2.22\,bpm, achieving accuracy comparable to controlled static sensing while remaining robust under unconstrained robot--human configurations.
Project page: \url{https://yuxuanhu9.github.io/ActiveVital/}.
}

\correspondence{\email{jianfei.yang@ntu.edu.sg}}

\begin{document}

\maketitle

\begin{figure*}
    \centering
    \includegraphics[width=\linewidth]{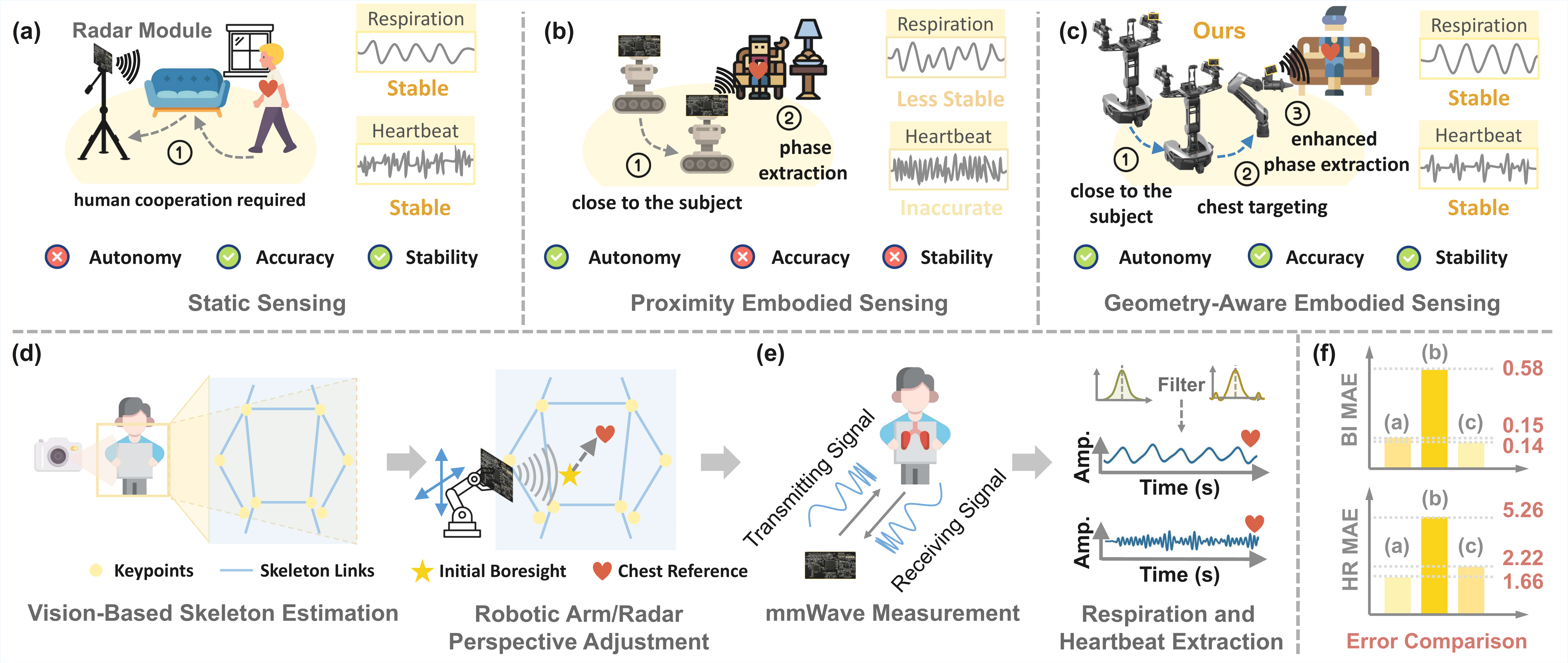}
    \caption{
    \textbf{Geometry-Aware Embodied mmWave Vital Signs Sensing (ActiveVital).}
    (a) Static Sensing relies on user alignment to maintain favorable observation geometry.
    (b) Proximity Embodied Sensing improves signal strength by reducing distance but does not regulate anatomical alignment.
    (c) Geometry-Aware Embodied Sensing actively controls sensor pose to achieve chest-centric geometric alignment and stable measurement.
    (d) Vision-based chest localization and geometric reference estimation.
    (e) Closed-loop radar alignment and mmWave-based respiration and heartbeat extraction.
    (f) Quantitative error comparison across sensing paradigms, demonstrating the benefit of active geometry regulation.
    }
    \label{fig:overview}
\end{figure*}

\section{Introduction}

Home robots are increasingly deployed for long-term in-home service and companionship. Beyond task execution, these robots are expected to support user well-being through health monitoring during daily interaction. Reliable respiration and heart rate monitoring is therefore critical for ensuring user safety and enabling health-aware interaction. However, obtaining these vital signs without physical contact remains challenging in unconstrained home environments. Millimeter-wave (mmWave) radar has been proven to be a promising solution, as it captures thoracic micro-motion contactlessly through phase variations~\citep{1.1.2, 1.1.4}. Beyond vital signs, wireless sensing has also advanced broader human perception tasks such as pose estimation, mesh reconstruction, and action recognition~\citep{3666122.3666944, Fan_2026_CVPR, chen2025xfimodalityinvariantfoundationmodel}.

Nevertheless, mmWave radar measures displacement only along the line of sight, meaning that only the radial component of respiration and heartbeat motion is observable. Consequently, geometric misalignment directly attenuates the measurable signal, particularly for heartbeat components with extremely small amplitudes. This geometry-induced observability constraint indicates that measurement quality is governed primarily by sensor--human relative pose rather than signal processing alone.

Most existing systems adopt a static sensing paradigm~\citep{1.1.3}, as illustrated in Fig.~\ref{fig:overview}(a), requiring users to remain aligned with the radar to maintain favorable observation geometry. Some studies deploy embodied sensing on mobile platforms~\citep{1.1.5, 1.1.6}, as shown in Fig.~\ref{fig:overview}(b), but they regulate only spatial proximity and improve measurement quality mainly through distance reduction. Consequently, current methods lack a closed-loop pose regulation mechanism that directly optimizes vital sign observability.

In domestic environments, where users may be sleeping, seated, or moving freely, pose and interaction behaviors are inherently uncertain, making sustained near-axial observation difficult. Consequently, observation geometry must be actively regulated by the robot rather than passively relying on user alignment. To address this challenge, we propose ActiveVital, a chest-centric embodied sensing framework (Fig.~\ref{fig:overview}(c)). Visual keypoints are used to estimate a chest reference, enabling the robot to regulate radar pose and achieve near-axial alignment with the thoracic surface within a closed-loop control framework. To ensure stable signal extraction under residual pose variations, a position-robust phase enhancement module is integrated for reliable respiration and heartbeat separation. The overall perception-to-alignment pipeline and quantitative comparison are illustrated in Fig.~\ref{fig:overview}(d)--(f).

The system is implemented on a mobile robot equipped with a 60\,GHz frequency-modulated continuous-wave (FMCW) radar and evaluated under three sensing modes: Static Sensing, Proximity Embodied Sensing, and Geometry-Aware Embodied Sensing. Experimental results show that closed-loop chest alignment significantly improves estimation accuracy, enabling stable non-contact vital sign monitoring in real domestic environments.

The main contributions are summarized as follows:

\begin{itemize}
    \item \textbf{Geometry-Regulated Embodied Sensing:}
    An embodied vital sign sensing paradigm that actively regulates radar observation geometry through feedback-driven pose alignment, enabling reliable sensing under unconstrained robot--human configurations.

    \item \textbf{Vision-Guided Pose Regulation with Robust Signal Extraction:}
    A vision-guided alignment pipeline steers the robot-mounted radar toward near-axial observation, while phase-based signal enhancement stabilizes respiration and heartbeat extraction under pose variations.

    \item \textbf{Cross-Paradigm and Cross-Baseline Validation:}
    Evaluation under three sensing paradigms (Static Sensing, Proximity Embodied Sensing, and Geometry-Regulated Embodied Sensing) with three baseline methods demonstrates consistent accuracy improvements over existing approaches.
\end{itemize}

\begin{figure*}
    \centering
    \includegraphics[width=\linewidth]{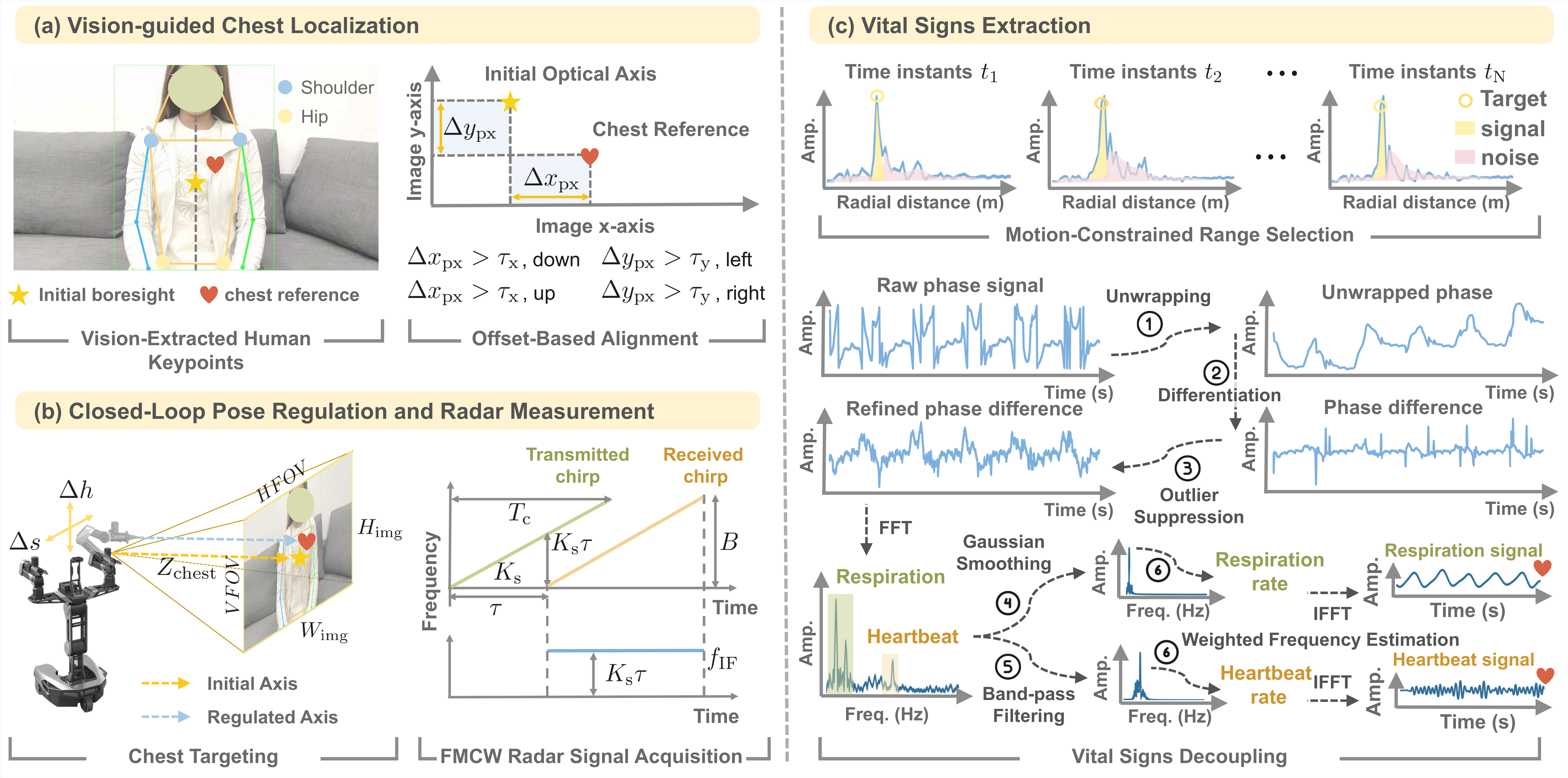}
    \caption{
    \textbf{Processing pipeline of the proposed Geometry-Aware Embodied Sensing framework.}
    (a) Vision-guided chest localization using human skeletal keypoints and image-space offset alignment to define a geometric reference.
    (b) Closed-loop pose regulation and FMCW radar signal acquisition, where the robotic arm actively adjusts sensor orientation to achieve chest-centric alignment.
    (c) Motion-constrained vital signs extraction, including range selection, phase refinement, spectral decoupling, and respiration/heartbeat rate estimation.
    }
    \label{fig:pipeline}
\end{figure*}

\section{Related Work}

This section reviews prior studies on non-contact vital signs monitoring from the perspective of perception--control coupling in robotic systems, highlighting the limitations of passive static observation and the need for geometry-aware embodied sensing as a controllable observation paradigm.

\subsection{Static Vital Signs Monitoring}

Non-contact vital signs monitoring has been explored using visible-light imaging, thermal sensing, acoustics, Wi-Fi, and radar~\citep{2.1.01,2.1.02,2.1.03,2.1.04,2.1.05}. Among these modalities, FMCW mmWave radar is widely adopted due to its sensitivity to thoracic micro-motion along the radar line of sight~\citep{2.1.07}. Most mmWave systems follow a static sensing paradigm in which the sensor is fixed and subjects are assumed to remain within favorable observation geometry. Under this assumption, research primarily concentrates on backend signal processing, including differential phase enhancement~\citep{2.12.2}, model-based separation~\citep{2.12.1}, and high-resolution spectral analysis~\citep{2.12.0}. These methods improve extraction performance when geometry is stable, but they treat observation geometry as externally imposed rather than actively controlled.
Recent multimodal approaches incorporate vision or thermal sensing to assist localization and motion compensation~\citep{2.13.1,2.13.2,2.13.3,2.13.4,2.13.5}. However, these systems still rely on passively configured sensing axes. Prior studies further indicate that misalignment between the radar probing vector and the effective thoracic surface reduces the radial projection of cardiac-induced micro-motion~\citep{2.12.1}, which degrades heartbeat-related phase observability.

\subsection{Proximity Embodied Vital Signs Sensing}

To alleviate fixed-view limitations, several studies embed mmWave radar into mobile robotic platforms, leveraging mobility to achieve spatial proximity and enhance echo amplitude~\citep{1.1.6, hu2026wavemanmmwavebasedroomscalehuman, lu2026emfallembodiedmmwavesensing}. However, existing approaches primarily emphasize distance reduction rather than structure-aware geometric alignment. The coupling between chest anatomy and radar observation geometry is rarely modeled explicitly, and observation pose is seldom treated as a controllable variable. Pose-aware systems that introduce directional adjustment mechanisms~\citep{1.1.5, 2.2.1} typically rely on heuristic rules without explicitly incorporating anatomical constraints.
In summary, existing work either optimizes signal extraction under fixed geometry or exploits mobility for proximity without achieving explicit regulation of radar line-of-sight alignment with the thoracic surface. A sensing paradigm that integrates anatomical priors with mmWave geometric constraints and treats observation geometry as a controllable variable remains largely unexplored. To address this limitation, we propose ActiveVital, which actively constructs near-axial alignment through vision-guided embodied alignment and combines it with a geometry-robust phase refinement module for stable respiration and heartbeat estimation under posture variation.

\section{Method}
\label{sec:method}

ActiveVital resolves the geometry-induced observability constraint through a three-stage pipeline integrating vision-guided pose regulation and phase-domain enhancement (Fig.~\ref{fig:pipeline}).
First, visual keypoints define a chest-centered reference in the image domain.
Second, this reference drives closed-loop pose control to align the radar line-of-sight with the thoracic surface normal.
Third, differential phase enhancement stabilizes signal extraction and separates respiration and heartbeat components.

\subsection{Chest Localization}
As illustrated in Fig.~\ref{fig:pipeline}(a), posture variations in non-cooperative scenarios can misalign the radar line-of-sight with the chest surface, reducing cardiopulmonary motion observability. To enable active pose regulation under such conditions, a chest-centered reference is defined in the image domain. We employ ViTPose~\citep{3.0.1} to detect four torso keypoints:
\begin{equation}
\label{eq:keypoints}
P_{\mathrm{ls}},\; P_{\mathrm{rs}},\; P_{\mathrm{lh}},\; P_{\mathrm{rh}} \in \mathbb{R}^2,
\end{equation}
where each keypoint is written as $P_i = [P_{x,i},\,P_{y,i}]^\top$ in an image coordinate system with the $x$-axis pointing rightward and the $y$-axis upward.
The shoulder and hip midpoints are computed as
\begin{equation}
\label{eq:midpoints}
P_{\mathrm{sc}} = \frac{P_{\mathrm{ls}} + P_{\mathrm{rs}}}{2}, \quad
P_{\mathrm{hc}} = \frac{P_{\mathrm{lh}} + P_{\mathrm{rh}}}{2}.
\end{equation}

The vertical coordinate of the chest reference is interpolated along the torso axis as
\begin{equation}
\label{eq:yref}
P_{y,\mathrm{ref}} = \alpha P_{y,\mathrm{sc}} + (1-\alpha) P_{y,\mathrm{hc}},
\end{equation}
where $\alpha \in (0,1)$. We set $\alpha = 0.75$ to bias the sensing anchor toward the upper chest region.
To compensate for lateral anatomical asymmetry, a horizontal offset proportional to shoulder width is introduced. Define
\begin{equation}
\label{eq:shoulderwidth}
W_{\mathrm{shoulder}} = \left| P_{x,\mathrm{ls}} - P_{x,\mathrm{rs}} \right|,
\end{equation}
and the horizontal coordinate is defined as
\begin{equation}
\label{eq:xref}
P_{x,\mathrm{ref}} = P_{x,\mathrm{sc}} + \beta W_{\mathrm{shoulder}},
\end{equation}
where $\beta = 0.1$ shifts the reference toward the cardiac-adjacent region.
The final chest sensing anchor is given by
\begin{equation}
\label{eq:pref}
P_{\mathrm{ref}} = \left[ P_{x,\mathrm{ref}},\; P_{y,\mathrm{ref}} \right]^\top.
\end{equation}

\subsection{Chest-Centric Observation Construction}

\subsubsection{Closed-Loop Error-Driven Pose Regulation}

As illustrated in Fig.~\ref{fig:pipeline}(b), the robot regulates its sensor pose so that the radar probing axis intersects the chest reference point $P_{\mathrm{ref}}$. Pixel deviations are converted into motion commands within a closed-loop control scheme.
Let the image dimensions be $W_{\mathrm{img}}$ and $H_{\mathrm{img}}$. The image center is defined as
\begin{equation}
\label{eq:image_center}
L_x = \frac{W_{\mathrm{img}}}{2}, \quad
L_y = \frac{H_{\mathrm{img}}}{2}.
\end{equation}

Let $P_{\mathrm{ref}} = [x,\, y]^\top$. The pixel error vector is
\begin{equation}
\label{eq:error_vector}
\mathbf{E} = [e_x,\, e_y]^\top,
\quad
e_x = x - L_x,\;
e_y = y - L_y.
\end{equation}

To suppress oscillatory motion caused by estimation noise, tolerance thresholds $\delta_x, \delta_y \in \mathbb{R}_+$ are introduced. When $|e_x| \le \delta_x$ and $|e_y| \le \delta_y$, no motion is executed. The control output is defined as $\mathbf{A} = [a_{\mathrm{v}},\, a_{\mathrm{h}}]^\top$,
where
\begin{equation}
\label{eq:control_law}
a_{\mathrm{v}} =
\begin{cases}
1, & e_y > \delta_y, \\
-1, & e_y < -\delta_y, \\
0, & \text{otherwise},
\end{cases}
\quad
a_{\mathrm{h}} =
\begin{cases}
1, & e_x > \delta_x, \\
-1, & e_x < -\delta_x, \\
0, & \text{otherwise}.
\end{cases}
\end{equation}

Here, $a_{\mathrm{h}} > 0$ and $a_{\mathrm{v}} > 0$ correspond to rightward and upward sensor motion in the robot coordinate frame, respectively.
The binary control vector $\mathbf{A}$ determines motion direction along the horizontal and vertical axes, while the motion magnitude is scaled according to the projected metric displacement derived from $\mathbf{E}$.
Alignment is established before each measurement, after which the seated subject stays still over a short window so the radar holds a fixed chest-centric pose.

\subsubsection{Image-to-Robot Spatial Mapping}

Pixel-level motion commands are converted into metric displacements using depth information and camera intrinsics.
Let $D$ denote the stereo depth map. The target distance is estimated within a neighborhood $\mathcal{N}(P_{\mathrm{ref}})$ as
\begin{equation}
\label{eq:depth_est}
D_{\mathrm{target}} =
\mathrm{median}
\{ D(u,v) \mid (u,v) \in \mathcal{N}(P_{\mathrm{ref}}) \}.
\end{equation}

Using the pinhole camera model with focal lengths $f_x$ and $f_y$, the spatial compensation corresponding to $\mathbf{E}$ is approximated by
\begin{equation}
\label{eq:metric_mapping}
\Delta H = a_{\mathrm{v}} \frac{|e_y|\, D_{\mathrm{target}}}{f_y},
\quad
\Delta L = a_{\mathrm{h}} \frac{|e_x|\, D_{\mathrm{target}}}{f_x}.
\end{equation}

Depth is regulated by $\Delta D = D_{\mathrm{target}} - D_{\mathrm{ref}}$, where $D_{\mathrm{ref}}$ is the desired sensing distance.
The resulting command $\mathbf{u} = [\Delta L,\, \Delta H,\, \Delta D]$ drives radar-chest alignment.

\subsection{Geometry-Constrained Phase Processing}
\subsubsection{Phase Modeling}

Under chest-centric alignment, radar phase variation is governed by radial cardiopulmonary displacement, modeled as
\begin{equation}
R(t)=R_0+\Delta R(t), \quad \Delta R(t)=d(t)\cos\theta,
\end{equation}
where $d(t)$ is the true chest displacement and $\theta$ is the angle between the radar line-of-sight and the chest normal. Active alignment drives $\theta \to 0$, maximizing radial observability.
\begin{equation}
\label{eq:tx}
S_{\mathrm{T}}(t)
=
A_{\mathrm{T}} \exp\!\left( j \left( 2\pi f_0 t + \pi K_s t^2 \right) \right),
\end{equation}
where $f_0$ is the carrier frequency and $K_{\mathrm{s}}$ is the chirp slope.
The received signal reflected from the chest is
\begin{equation}
\label{eq:rx}
S_{\mathrm{R}}(t)
=
A_{\mathrm{R}}
\exp\!\left(
j \left[
2\pi f_0 (t-\tau)
+ \pi K_s (t-\tau)^2
+ 2\pi f_{\mathrm{d}} (t-\tau)
\right]
\right),
\end{equation}
where $\tau(t)=2R(t)/c$ and $f_{\mathrm{d}}$ capture range modulation and micro-Doppler induced by chest motion.
After mixing and low-pass filtering, the intermediate-frequency (IF) signal is obtained, whose phase encodes radial displacement.

Under the Geometrical Theory of Diffraction (GTD) interpretation, the chest echo can be approximated by a finite set of dominant scattering centers. For the $i$-th scatterer with distance $R_i(t)$ and delay $\tau_i(t)=2R_i(t)/c$, the corresponding IF phase term is approximated by
\begin{equation}
\label{eq:if_phase}
\phi_{\mathrm{if}}^{(i)}(t)
\approx
2\pi f_0 \tau_i
+
2\pi K_{\mathrm{s}} \tau_i t
-
2\pi f_{\mathrm{d}}^{(i)} t.
\end{equation}

A range FFT is applied to obtain the range profile, and the chest bin $j_{\mathrm{trgt}}$ is selected by amplitude-based detection with temporal continuity constraint. The slow-time phase sequence is then constructed as
\begin{equation}
\label{eq:phase_seq}
\phi(m)
=
\arg\!\left( S_{\mathrm{IF}}(j_{\mathrm{trgt}}, m) \right),
\quad m=1,\dots,M.
\end{equation}

\subsubsection{Differential Phase Enhancement}

The raw phase sequence $\phi(m)$ is first unwrapped to obtain a continuous trajectory $\phi_{\mathrm{unwrap}}(m)$.

To suppress low-frequency drift and enhance periodic micro-motion, first-order temporal differencing is performed:
\begin{equation}
\label{eq:diff}
\Delta \phi_{\mathrm{unwrap}}(m)
=
\phi_{\mathrm{unwrap}}(m)
-
\phi_{\mathrm{unwrap}}(m-1).
\end{equation}

To mitigate impulsive noise, a Hampel filter is applied within the window
\begin{equation}
\label{eq:window}
\Omega_m = \{ m-r,\ldots,m+r \}.
\end{equation}

Within this window, the local median and deviation are computed as
\begin{equation}
\label{eq:mad}
\mathrm{med}(m)
=
\mathrm{median}\{\Delta \phi_{\mathrm{unwrap}}(i)\},
\end{equation}

\begin{equation}
\label{eq:mad2}
\mathrm{MAD}(m)
=
\mathrm{median}
\left|
\Delta \phi_{\mathrm{unwrap}}(i)
-
\mathrm{med}(m)
\right|.
\end{equation}

The robust scale estimate is
\begin{equation}
\label{eq:sigma}
\sigma(m) = \kappa \,\mathrm{MAD}(m),
\end{equation}
where $\kappa = 1.4826$.

Samples satisfying
\begin{equation}
\label{eq:hampel}
\left|
\Delta \phi_{\mathrm{unwrap}}(m)
-
\mathrm{med}(m)
\right|
>
\tau \sigma(m)
\end{equation}
are replaced by $\mathrm{med}(m)$ to suppress outliers.

The refined phase sequence is denoted as $\widetilde{\phi}(m)$.

\subsubsection{Respiration and Heartbeat Rate Estimation}

The magnitude spectrum of the refined phase sequence is computed as
\begin{equation}
\label{eq:spectrum}
P(f)
=
\left|
\mathrm{FFT}\{\widetilde{\phi}(m)\}
\right|.
\end{equation}

Respiration and heartbeat components are separated via Gaussian smoothing and FIR band-pass filtering, yielding $P_{\mathrm{resp}}(f)$ and $P_{\mathrm{heart}}(f)$.

The heart rate is estimated via energy-weighted averaging within the cardiac band:
\begin{equation}
\label{eq:hr_freq}
f_{\mathrm{heart}}
=
\frac{\sum_k f_k P_{\mathrm{heart}}(f_k)}
{\sum_k P_{\mathrm{heart}}(f_k)}.
\end{equation}

The final heart rate is $\mathrm{HR} = 60 f_{\mathrm{heart}}$.

For respiration, if $K$ peaks are detected between $t_1^{\mathrm{resp}}$ and $t_K^{\mathrm{resp}}$, the breathing rate is
\begin{equation}
\label{eq:br}
\mathrm{BR}
=
\frac{K-1}{t_K^{\mathrm{resp}} - t_1^{\mathrm{resp}}}
\times 60.
\end{equation}

\begin{figure}[t]
    \centering
    \includegraphics[width=\linewidth]{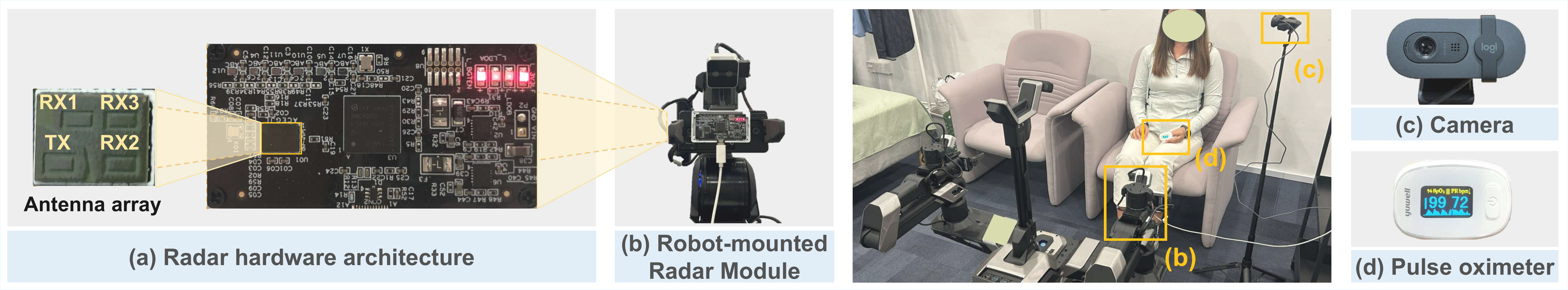}
    \caption{
    \textbf{Experimental hardware configuration and scene setup.}
    (a) 60\,GHz FMCW radar hardware, including antenna array and radar board.
    (b) Robot-mounted radar with onboard RGB-D for chest localization.
    (c) External RGB camera for the respiration-rate reference.
    (d) Fingertip pulse oximeter providing reference HR.
    The robot-mounted radar actively regulates its pose toward the seated participant for embodied vital signs sensing.
    }
    \label{fig:3}
\end{figure}

\begin{figure}[t]
    \centering
    \includegraphics[width=\linewidth]{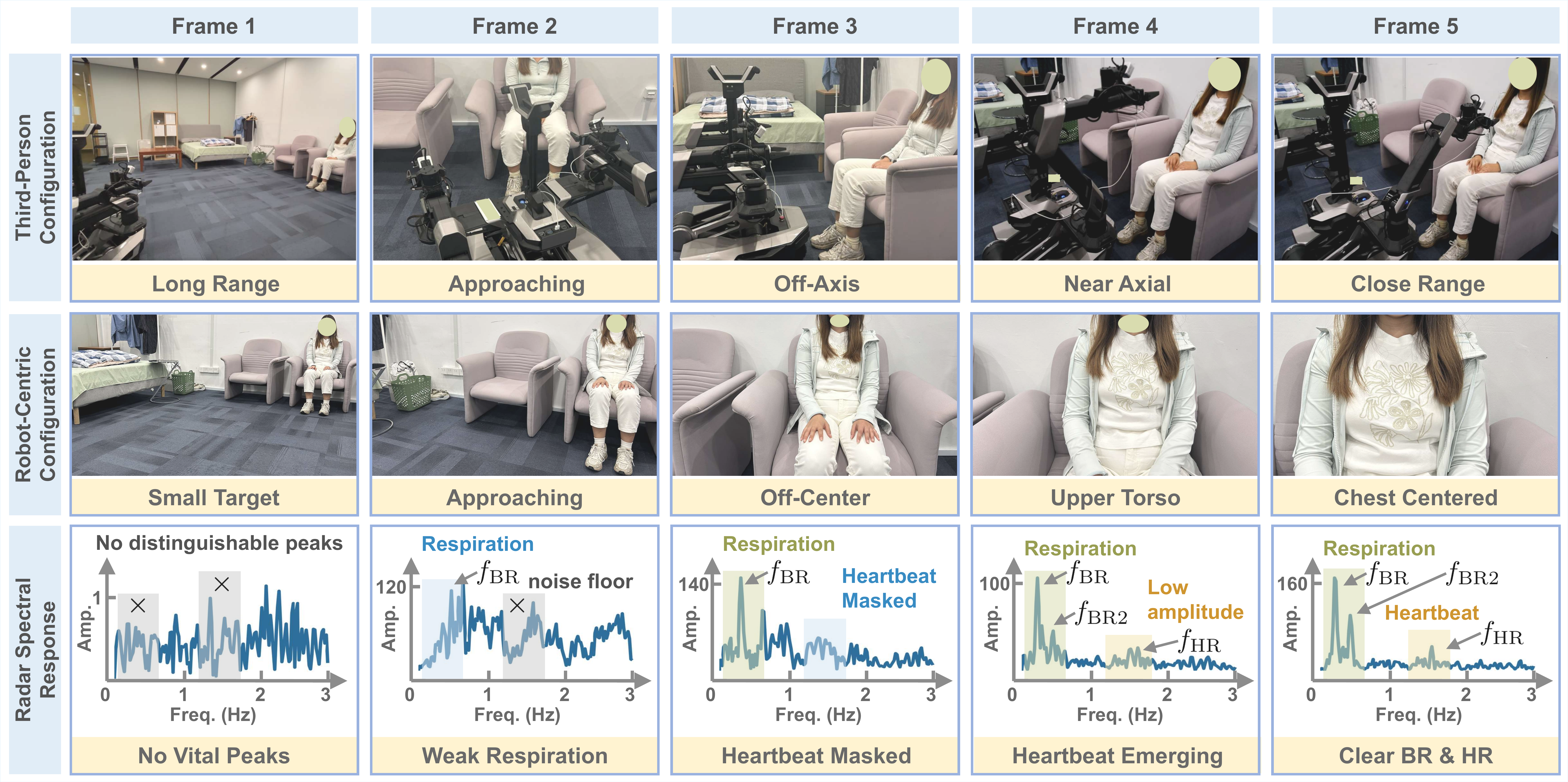}
    \caption{
    \textbf{Progressive evolution of observation geometry and corresponding spectral responses.}
    Upper panels show representative physical configurations from long-range and off-axis conditions to chest-centered alignment.
    Lower panels present the associated phase-derived spectra, illustrating the transition from indistinguishable signals to clearly separable respiration and heartbeat components.
    }
    \label{fig:4}
\end{figure}

\section{Experiment}

\subsection{Experimental Setup}

The proposed ActiveVital system is implemented on a Galaxea R1 Lite mobile robotic platform integrating millimeter-wave radar and RGB-D vision modules, as illustrated in Fig.~\ref{fig:3}. The radar module is rigidly mounted on the robot arm to enable active pose regulation, while an RGB camera provides visual keypoints for chest localization.

Specifically, the system incorporates a 60\,GHz FMCW radar with a 1T3R antenna configuration (Fig.~\ref{fig:3}(a)(b)). The radar operates over 57.5--63.5\,GHz with a chirp slope of 91.5\,MHz/$\mu$s and a sampling rate of 2\,MHz. Each chirp lasts 71.8\,$\mu$s with a frame period of 50\,ms, yielding a range resolution of approximately 0.0228\,m. Raw ADC data are streamed to a host computer for signal processing.

An onboard RGB-D camera (Fig.~\ref{fig:3}(b)) extracts human skeletal keypoints for geometric reference construction. Pixel deviations between the radar optical axis and the chest reference are converted into metric displacements through the image-to-robot mapping described in Section~\ref{sec:method}, enabling closed-loop regulation of the sensing pose.

All experiments are conducted in a 6.6\,m $\times$ 6.3\,m $\times$ 3.5\,m indoor environment under controlled lighting conditions. Participants remain seated in a natural posture without explicit alignment instructions. Heart rate ground truth is synchronously recorded using a fingertip pulse oximeter (Fig.~\ref{fig:3}(d)), while respiration rate is derived from chest motion measured by the RGB camera (Fig.~\ref{fig:3}(c)) as an external reference for evaluation.

At the beginning of each trial, the robot is positioned approximately 1\,m in front of the subject, with the chest located within the field of view of the arm-mounted camera. Chest localization is then performed using ViTPose~\citep{3.0.1}, and chest-centric alignment is achieved within a few seconds.

To illustrate the impact of observation geometry, Fig.~\ref{fig:4} presents representative configurations from long-range and off-axis conditions to chest-centered alignment. The corresponding spectral responses demonstrate the progressive emergence of respiration and heartbeat components as geometric alignment improves.

\begin{figure}
    \centering
    \includegraphics[width=0.6\linewidth]{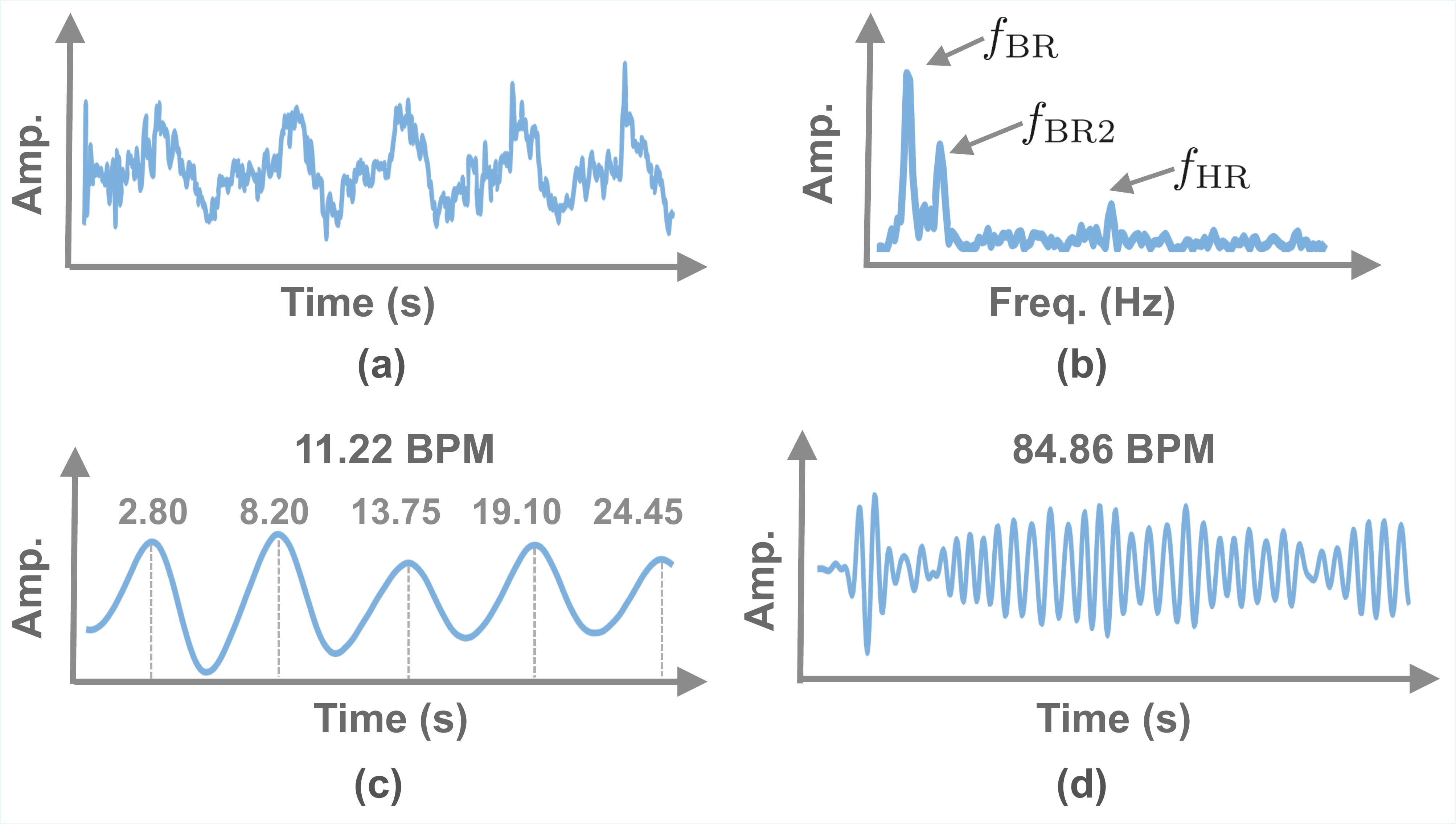}
    \caption{
    Radar-measured phase signal representations under ActiveVital.
    (a) Slow-time unwrapped phase after phase refinement.
    (b) Corresponding amplitude spectrum showing distinct respiration fundamental ($f_{\mathrm{BR}}$), respiration harmonic ($f_{\mathrm{BR2}}$), and heartbeat ($f_{\mathrm{HR}}$) components.
    (c) Extracted respiration waveform and estimated breathing rate.
    (d) Extracted heartbeat waveform and estimated heart rate.
    }
    \label{fig:5}
\end{figure}

\begin{figure}
    \centering
    \includegraphics[width=0.6\linewidth]{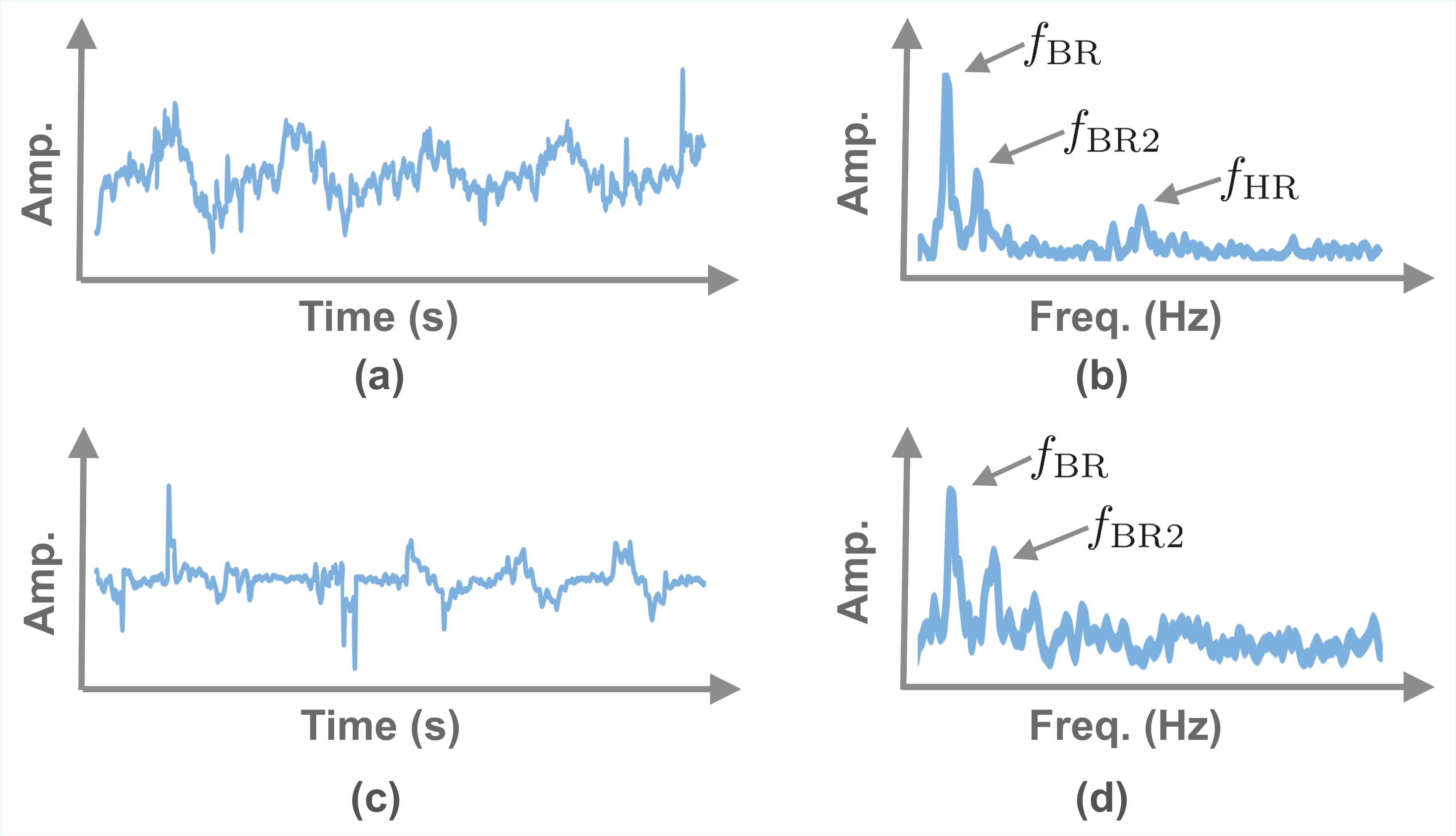}
    \caption{
    Radar-measured phase signal representations under Static and Proximity Embodied Sensing.
    (a)(c) Slow-time phase sequences.
    (b)(d) Corresponding amplitude spectra.
    The first row corresponds to Static Sensing, and the second to Proximity Embodied Sensing.
    }
    \label{fig:6}
\end{figure}

\subsection{Evaluation Across Sensing Paradigms}

\subsubsection{Geometry-Dependent Observability}

We compare three sensing paradigms under identical hardware configuration and signal processing pipeline: Static Sensing, Proximity Embodied Sensing, and the proposed Geometry-Aware Embodied Sensing.
They differ only in geometry. In Static sensing, the radar is fixed and the seated subject faces it while manually keeping near-axial alignment. In Proximity sensing, the robot uses the same keypoints to approach and shorten the distance but does not align the radar to the chest~\citep{1.1.5, 1.1.6}. In the Geometry-aware setting (ActiveVital), the robot aligns the radar with the chest to reach near-normal incidence.
Fig.~\ref{fig:5} shows the refined slow-time phase sequence measured from the radar signal and its corresponding spectrum under ActiveVital. The respiration fundamental, harmonic, and heartbeat peaks are clearly separated, enabling stable vital signs extraction.
Fig.~\ref{fig:6} presents the corresponding results under Static and Proximity Embodied Sensing. Static Sensing preserves distinguishable respiration and heartbeat peaks because subjects are deliberately positioned to maintain near-axial alignment with the radar line of sight. In contrast, under Proximity Embodied Sensing, misalignment reduces the radial projection of cardiac micro-motion. The attenuated heartbeat component falls close to the noise floor, making the spectral peak indistinguishable from background noise and leading to unstable estimation.
These results indicate that deviation from optimal geometric alignment reduces the radial projection amplitude of cardiac micro-motion, thereby degrading its spectral observability.

\begin{figure}
    \centering
    \includegraphics[width=0.5\linewidth]{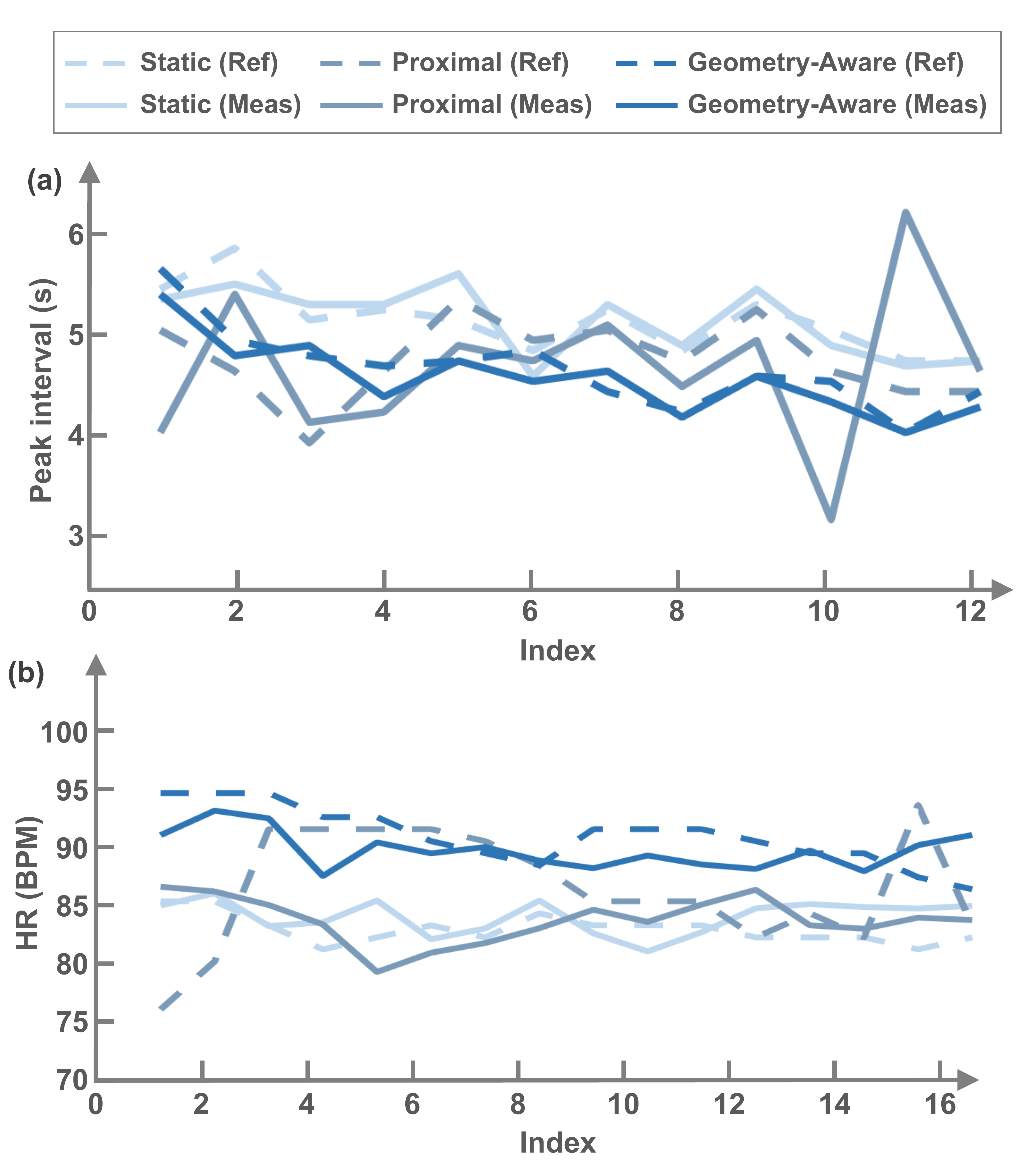}
    \caption{
    Estimation results under three sensing paradigms using the same signal processing method.
    (a) BI estimation across 12 peak intervals under Static Sensing, Proximity Embodied Sensing, and ActiveVital, compared with reference measurements.
    (b) HR estimation across 16 sliding-window segments under the same paradigms, with corresponding reference values.
    }
    \label{fig:7}
\end{figure}

\begin{figure}
    \centering
    \includegraphics[width=0.5\linewidth]{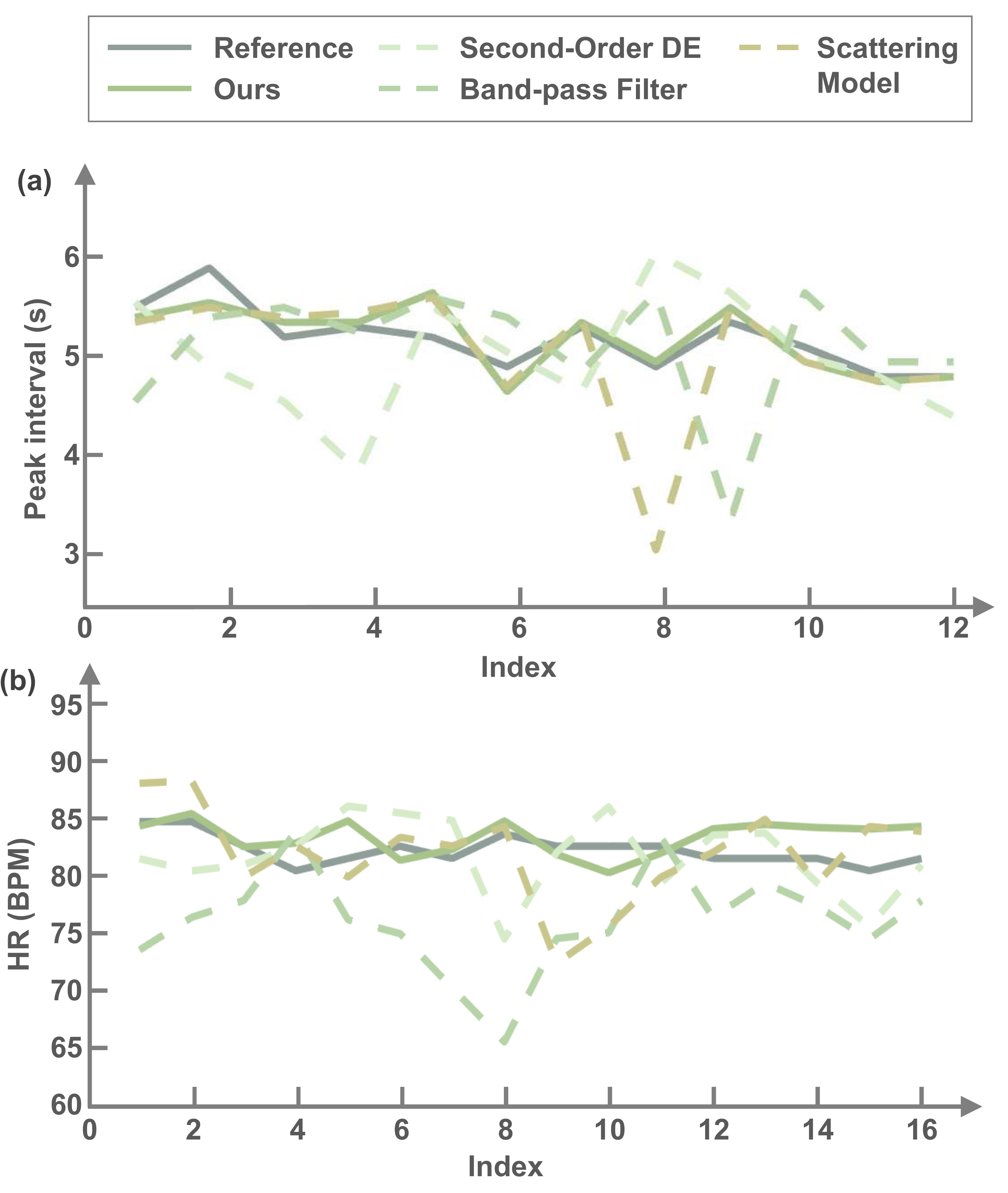}
    \caption{
    Algorithmic performance under Static Sensing.
    (a) Breath interval estimation across 12 peak intervals.
    (b) Heart rate estimation across 16 time segments.
    The proposed sequential phase refinement method is compared with three baseline approaches against reference measurements.
    }
    \label{fig:8}
\end{figure}

\subsubsection{Geometry-Dependent Estimation Accuracy}

Measurement results of breath interval (BI) and heart rate (HR) under the three sensing paradigms are shown in Fig.~\ref{fig:7}, where estimated values are compared against reference measurements. Quantitative accuracy is further evaluated using mean absolute error (MAE) and root mean square error (RMSE), as summarized in Table~\ref{tab:cross_configuration}.
BI is computed from 75\,s of data, yielding 12 peak-to-peak intervals. HR is estimated from 50\,s of data using a sliding window (20\,s initialization, 2\,s update), resulting in 16 HR estimates.
For BI, Static and Geometry-aware Sensing achieve comparable accuracy (MAE 0.15\,s vs.\ 0.14\,s; RMSE 0.19\,s vs.\ 0.17\,s), while Proximity Sensing exhibits significantly larger errors (MAE 0.58\,s; RMSE 0.78\,s).
For HR, Static yields the lowest error (MAE 1.66\,bpm; RMSE 1.96\,bpm). Geometry-aware Sensing achieves comparable performance (MAE 2.22\,bpm; RMSE 2.59\,bpm), substantially outperforming Proximity Sensing (MAE 5.26\,bpm; RMSE 6.59\,bpm).

Although Static Sensing achieves the lowest HR error under manually enforced near-axial alignment, this configuration requires explicit subject cooperation to maintain favorable geometry.
In contrast, ActiveVital autonomously regulates the radar--thorax alignment, constructing comparable observation geometry without requiring posture constraints.
These results demonstrate that preserving near-axial alignment maintains the radial projection amplitude of cardiac micro-motion, thereby stabilizing heartbeat estimation under realistic posture variation.

\begin{table}[t]
\setlength{\tabcolsep}{11pt}
\centering
\caption{Quantitative comparison of BI and HR under different sensing paradigms using the proposed method.}
\label{tab:cross_configuration}
\begin{tabular}{lcccc}
\toprule
\multirow{2}{*}{\textbf{Paradigm}}
& \multicolumn{2}{c}{\textbf{BI (s)}}
& \multicolumn{2}{c}{\textbf{HR (bpm)}} \\
\cmidrule(lr){2-3} \cmidrule(lr){4-5}
& MAE & RMSE
& MAE & RMSE \\
\midrule
\textbf{Static}
& 0.15 & 0.19
& \cellcolor{morandipink}1.66 & \cellcolor{morandipink}1.96 \\

\textbf{Proximity}
& 0.58 & 0.78
& 5.26 & 6.59 \\

\textbf{ActiveVital}
& \cellcolor{morandipink}0.14 & \cellcolor{morandipink}0.17
& \cellcolor{morandiyellow}2.22 & \cellcolor{morandiyellow}2.59 \\

\bottomrule
\end{tabular}
\end{table}

\begin{table*}[t]
\setlength{\tabcolsep}{7pt}
\centering
\caption{Cross-algorithm comparison under three sensing paradigms (MAE and RMSE for BI and HR).}
\label{tab:cross_algorithm_wide}
\resizebox{\linewidth}{!}{%
\begin{tabular}{lcccccccccccc}
\toprule
\multirow{3}{*}{\textbf{Algorithm}}
& \multicolumn{4}{c}{\textbf{Static}}
& \multicolumn{4}{c}{\textbf{Proximal}}
& \multicolumn{4}{c}{\textbf{Geometry-aware}} \\
\cmidrule(lr){2-5} \cmidrule(lr){6-9} \cmidrule(lr){10-13}
& \multicolumn{2}{c}{BI (s)} & \multicolumn{2}{c}{HR (bpm)}
& \multicolumn{2}{c}{BI (s)} & \multicolumn{2}{c}{HR (bpm)}
& \multicolumn{2}{c}{BI (s)} & \multicolumn{2}{c}{HR (bpm)} \\
\cmidrule(lr){2-3} \cmidrule(lr){4-5}
\cmidrule(lr){6-7} \cmidrule(lr){8-9}
\cmidrule(lr){10-11} \cmidrule(lr){12-13}
& MAE & RMSE & MAE & RMSE
& MAE & RMSE & MAE & RMSE
& MAE & RMSE & MAE & RMSE \\
\midrule

\makecell[l]{Second-Order DE \\ \citep{2.12.2}}
& 0.51 & 0.68 & 2.93 & 3.44
& 0.65 & 0.75 & 9.27 & 12.12
& 0.93 & 1.40 & 7.60 & 9.48 \\

\makecell[l]{Band-pass Filter \\ \citep{1.1.6, 2.12.0}}
& 0.56 & 0.75 & 6.29 & 7.36
& 0.87 & 1.18 & 13.59 & 15.07
& 0.30 & 0.37 & 14.48 & 15.88 \\

\makecell[l]{Scattering Model \\ \citep{2.12.1}}
& 0.31 & 0.57 & 2.81 & 3.60
& 0.60 & 0.89 & 11.82 & 13.67
& \cellcolor{morandipink}{0.12} & \cellcolor{morandipink}{0.16} & 6.78 & 8.55 \\

\textbf{ActiveVital (Ours)}
& \cellcolor{morandipink}{0.15} & \cellcolor{morandipink}{0.19} & \cellcolor{morandipink}{1.65} & \cellcolor{morandipink}{1.96}
& \cellcolor{morandipink}{0.58} & \cellcolor{morandipink}{0.78} & \cellcolor{morandipink}{5.26} & \cellcolor{morandipink}{6.59}
& \cellcolor{morandiyellow}0.14 & \cellcolor{morandiyellow}0.17 & \cellcolor{morandipink}{2.22} & \cellcolor{morandipink}{2.59} \\

\bottomrule
\end{tabular}}
\end{table*}

\begin{figure}
    \centering
    \includegraphics[width=0.6\linewidth]{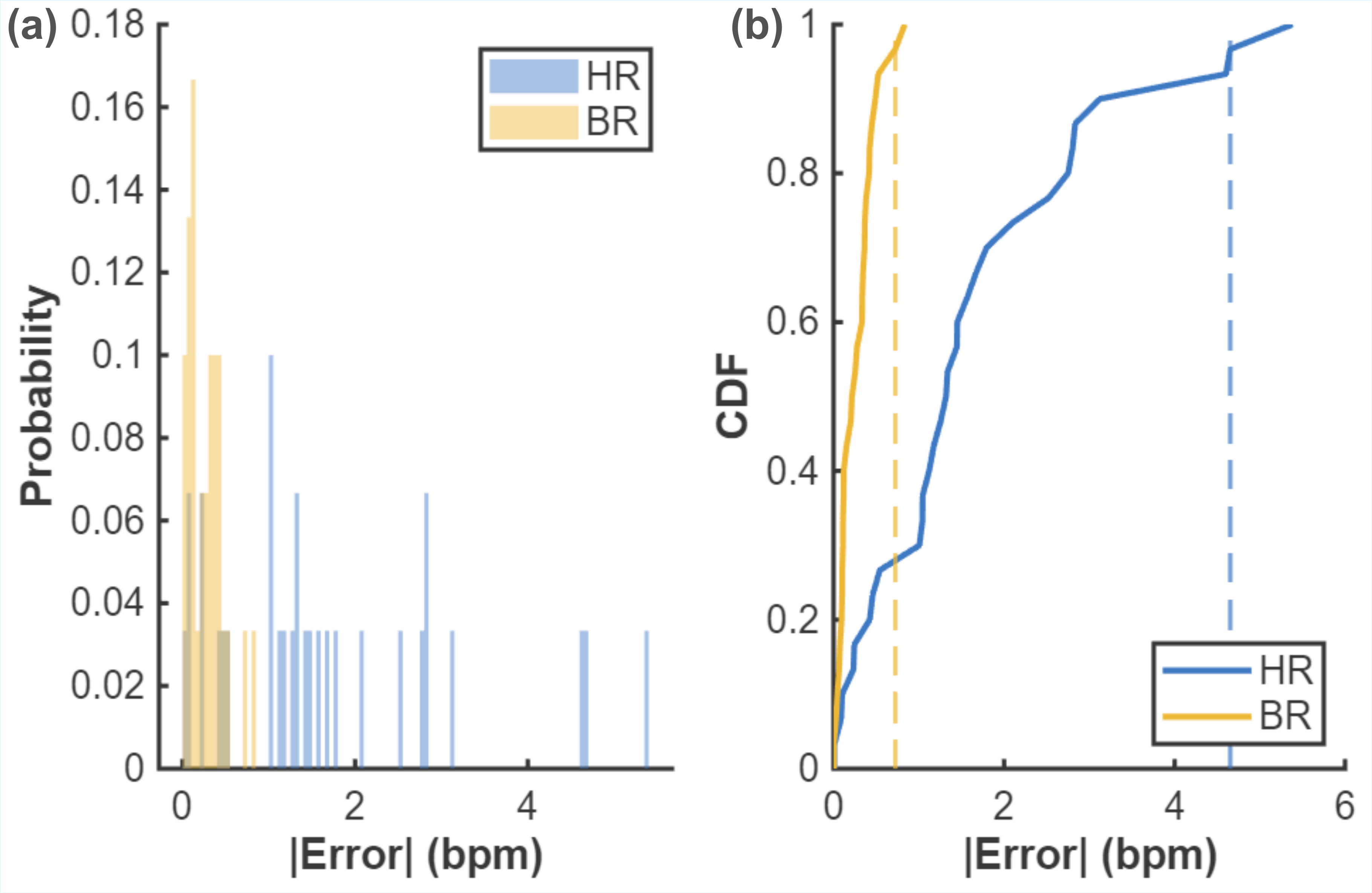}
    \caption{
    Distribution of absolute estimation errors for HR and BR.
    (a) Histogram of absolute errors.
    (b) Cumulative distribution function (CDF).
    Dashed vertical lines indicate the 95th percentile error thresholds.
    }
    \label{fig:9}
\end{figure}

\begin{table}[t]
\setlength{\tabcolsep}{11pt}
\centering
\renewcommand{\arraystretch}{1.25}
\caption{Statistical error metrics for HR and BR estimation over 30 independent recordings (25\,s each).}
\label{tab:error_stats}
\begin{tabular}{lccc}
\toprule
Metric & MAE (bpm) & RMSE (bpm) & P95 (bpm) \\
\midrule
BR & 0.26 & 0.33 & 0.72 \\
HR & 1.67 & 2.16 & 4.65 \\
\bottomrule
\end{tabular}
\end{table}

\subsection{Algorithm Performance Across Sensing Paradigms}

To analyze how different signal processing algorithms perform under varying sensing paradigms, we compare four methods across Static Sensing, Proximity Embodied Sensing, and ActiveVital.
While the previous subsection examined the effect of geometry on observability and estimation accuracy, this subsection investigates the interaction between sensing geometry and algorithm design.
Fig.~\ref{fig:8} presents temporal comparisons under \textit{Static Sensing}. Aggregated MAE and RMSE results across all three sensing paradigms are summarized in Table~\ref{tab:cross_algorithm_wide}.

Under \textit{Static Sensing}, the proposed ActiveVital method achieves the lowest MAE for both BI (0.15\,s) and HR (1.65\,bpm), outperforming all baseline methods.
When switching to \textit{Proximity Embodied Sensing}, errors increase for all algorithms due to degraded observation geometry. However, the proposed method exhibits smaller performance degradation, particularly in HR estimation.
Under \textit{Geometry-aware}, BI estimation remains competitive across methods, while the proposed approach achieves the lowest HR error (2.22\,bpm), indicating stronger preservation of weak cardiac-induced phase components.

Across sensing paradigms, all methods capture overall rhythmic trends. However, baseline approaches exhibit more pronounced error amplification as geometric conditions deteriorate. For example, the HR MAE of Baseline2 increases from 6.29\,bpm under \textit{Static} to 13.59\,bpm under \textit{Proximity}, whereas the proposed method increases from 1.65\,bpm to 5.26\,bpm. This cross-paradigm comparison demonstrates that the proposed algorithm maintains superior stability under geometry-induced observability variation.

\subsection{Statistical Evaluation of BR and HR Estimation}

To assess consistency, we analyzed 30 independent 25\,s recordings, each yielding one breathing rate (BR) and HR estimate against the reference. Fig.~\ref{fig:9} shows the absolute-error distribution (histogram and CDF) and Table~\ref{tab:error_stats} summarizes MAE, RMSE, and P95. BR errors are tightly concentrated (MAE 0.26\,bpm, P95 0.72\,bpm), while HR shows broader but bounded dispersion (MAE 1.67\,bpm, P95 4.65\,bpm) due to the weaker, lower-SNR cardiac micro-motion.

\section{Conclusion}

This work introduced ActiveVital, an embodied sensing framework that regulates radar--chest alignment through vision-guided pose control and phase-domain enhancement to maximize radial cardiopulmonary observability. Experiments across three sensing paradigms show that cardiac observability is geometry-dependent, and ActiveVital attains accuracy comparable to static sensing without active subject cooperation while clearly surpassing proximity-based sensing, with the largest gains in heart rate where cardiac signals are weak. These results establish sensing pose as a controllable variable for robust non-contact physiological monitoring. As an opportunistic function of a multipurpose home robot, ActiveVital provides contactless monitoring at a comfortable standoff without wearables or pre-installed infrastructure.

\section*{Acknowledgments}
This work is jointly supported by MOE Singapore Tier 1 Grant RG83/25, RS36/24 and a Start-up Grant from Nanyang Technological University.

\bibliographystyle{assets/plainnat}
\bibliography{IEEEabrv,mybib}

\begin{thebibliography}{26}
\providecommand{\natexlab}[1]{#1}
\providecommand{\url}[1]{\texttt{#1}}
\expandafter\ifx\csname urlstyle\endcsname\relax
  \providecommand{\doi}[1]{doi: #1}\else
  \providecommand{\doi}{doi: \begingroup \urlstyle{rm}\Url}\fi

\bibitem[Chen and Yang(2025)]{chen2025xfimodalityinvariantfoundationmodel}
Xinyan Chen and Jianfei Yang.
\newblock X-fi: A modality-invariant foundation model for multimodal human sensing.
\newblock \emph{arXiv preprint arXiv:2410.10167}, 2025.

\bibitem[Chian et~al.(2022)Chian, Wen, Wang, Hsu, and Wang]{2.13.2}
De-Ming Chian, Chao-Kai Wen, Chang-Jen Wang, Ming-Huan Hsu, and Fu-Kang Wang.
\newblock Vital signs identification system with doppler radars and thermal camera.
\newblock \emph{IEEE Transactions on Biomedical Circuits and Systems}, 16\penalty0 (1):\penalty0 153--167, 2022.

\bibitem[Cittadini et~al.(2024)Cittadini, Buonocore, Di~Castro, and Zollo]{1.1.5}
Roberto Cittadini, Luca~Rosario Buonocore, Mario Di~Castro, and Loredana Zollo.
\newblock Contactless respiration rate monitoring and human body pose detection for search and rescue robots.
\newblock In \emph{2024 10th IEEE RAS/EMBS International Conference for Biomedical Robotics and Biomechatronics (BioRob)}, pages 1119--1125, 2024.

\bibitem[Fan et~al.(2026)Fan, Zhou, Yang, Cui, Zhang, Xie, Yang, Lu, and Ding]{Fan_2026_CVPR}
Junqiao Fan, Yunjiao Zhou, Yizhuo Yang, Xinyuan Cui, Jiarui Zhang, Lihua Xie, Jianfei Yang, Chris~Xiaoxuan Lu, and Fangqiang Ding.
\newblock M4human: A large-scale multimodal mmwave radar benchmark for human mesh reconstruction.
\newblock In \emph{Proceedings of the IEEE/CVF Conference on Computer Vision and Pattern Recognition (CVPR)}, pages 42836--42846, 2026.

\bibitem[Gou et~al.(2026)Gou, Xiong, Lin, Hong, and Peng]{2.13.5}
Yingjie Gou, Yuyong Xiong, Wenjie Lin, Sicheng Hong, and Zhike Peng.
\newblock Vision-assisted chest wall localization for microwave vital sign monitoring.
\newblock \emph{IEEE Sensors Journal}, 26\penalty0 (2):\penalty0 2709--2723, 2026.

\bibitem[Hessler et~al.(2020)Hessler, Abouelenien, and Burzo]{2.1.02}
Christian Hessler, Mohamed Abouelenien, and Mihai Burzo.
\newblock A non-contact method for extracting heart and respiration rates.
\newblock In \emph{2020 17th Conference on Computer and Robot Vision (CRV)}, pages 1--8, 2020.

\bibitem[Hu et~al.(2024)Hu, Xia, and Xu]{2.12.1}
Yuxuan Hu, Zhaoyang Xia, and Feng Xu.
\newblock Investigating respiration-heartbeat separation through a multipoint scattering chest wall motion model: 60-ghz fmcw radar assessment.
\newblock \emph{IEEE Transactions on Instrumentation and Measurement}, 73:\penalty0 1--13, 2024.

\bibitem[Hu et~al.(2026)Hu, Zuo, Ma, Li, Xia, Xu, and Yang]{hu2026wavemanmmwavebasedroomscalehuman}
Yuxuan Hu, Kuangji Zuo, Boyu Ma, Shihao Li, Zhaoyang Xia, Feng Xu, and Jianfei Yang.
\newblock Waveman: mmwave-based room-scale human interaction perception for humanoid robots.
\newblock \emph{arXiv preprint arXiv:2601.07454}, 2026.

\bibitem[Huang et~al.(2022)Huang, Chen, Chai, Ehmke, Rupp, Dadabhoy, Feng, Li, Thomas, da~Silva, Boyer, and Traverso]{2.2.1}
Hen-Wei Huang, Jack Chen, Peter~R. Chai, Claas Ehmke, Philipp Rupp, Farah~Z. Dadabhoy, Annie Feng, Canchen Li, Akhil~J. Thomas, Marco da~Silva, Edward~W. Boyer, and Giovanni Traverso.
\newblock Mobile robotic platform for contactless vital sign monitoring.
\newblock \emph{Cyborg and Bionic Systems}, 2022:\penalty0 9780497, 2022.

\bibitem[Lee and Yang(2023)]{2.1.05}
In-Seong Lee and Jong-Ryul Yang.
\newblock Signal preprocessing for heartbeat detection using continuous-wave doppler radar.
\newblock \emph{IEEE Microwave and Wireless Technology Letters}, 33\penalty0 (4):\penalty0 479--482, 2023.

\bibitem[Li et~al.(2025)Li, Ayuknso, Yang, Wei, Zhi, Hua, and Zhao]{1.1.3}
Hongzhi Li, Ayukocha Gandhi~Bessemntoh Ayuknso, Nanyi Yang, Zeying Wei, Minghan Zhi, Qinglong Hua, and Bin Zhao.
\newblock Radar-based vital signal extraction via morphological component analysis.
\newblock \emph{IEEE Transactions on Microwave Theory and Techniques}, 73\penalty0 (11):\penalty0 9263--9278, 2025.

\bibitem[Liu et~al.(2018)Liu, Chen, Wang, Chen, Cheng, and Yang]{2.1.04}
Jian Liu, Yingying Chen, Yan Wang, Xu~Chen, Jerry Cheng, and Jie Yang.
\newblock Monitoring vital signs and postures during sleep using wifi signals.
\newblock \emph{IEEE Internet of Things Journal}, 5\penalty0 (3):\penalty0 2071--2084, 2018.

\bibitem[Lu et~al.(2026)Lu, Hu, Yuan, Zhou, Zuo, Lyu, Yuan, and Yang]{lu2026emfallembodiedmmwavesensing}
Yanshuo Lu, Yuxuan Hu, Shenghai Yuan, Xinyu Zhou, Kuangji Zuo, Bofan Lyu, XiChen Yuan, and Jianfei Yang.
\newblock Em-fall: Embodied mmwave sensing for day-and-night fall detection on humanoid robots.
\newblock \emph{arXiv preprint arXiv:2606.11109}, 2026.

\bibitem[Mercuri et~al.(2019)Mercuri, Lorato, Liu, Wieringa, Hoof, and Torfs]{1.1.2}
Marco Mercuri, Ilde~Rosa Lorato, Yao-Hong Liu, Fokko Wieringa, Chris~Van Hoof, and Tom Torfs.
\newblock Vital-sign monitoring and spatial tracking of multiple people using a contactless radar-based sensor.
\newblock \emph{Nature Electronics}, 2\penalty0 (6):\penalty0 252–262, June 2019.
\newblock ISSN 2520-1131.

\bibitem[Mirhosseini et~al.(2026)Mirhosseini, Alaee-Kerahroodi, Olk, Schroeder, and Mysore~R]{2.12.0}
Seyedeh~Fatemeh Mirhosseini, Mohammad Alaee-Kerahroodi, Andreas~E. Olk, Udo Schroeder, and Bhavani~Shankar Mysore~R.
\newblock In-car life detection and vital signs monitoring using mmwave radar sensors.
\newblock \emph{IEEE Transactions on Radar Systems}, 4:\penalty0 207--231, 2026.

\bibitem[Scebba et~al.(2021)Scebba, Da~Poian, and Karlen]{2.1.01}
Gaetano Scebba, Giulia Da~Poian, and Walter Karlen.
\newblock Multispectral video fusion for non-contact monitoring of respiratory rate and apnea.
\newblock \emph{IEEE Transactions on Biomedical Engineering}, 68\penalty0 (1):\penalty0 350--359, 2021.

\bibitem[Schroth et~al.(2024)Schroth, Eckrich, Kakouche, Fabian, von Stryk, Zoubir, and Muma]{1.1.6}
Christian~A. Schroth, Christian Eckrich, Ibrahim Kakouche, Stefan Fabian, Oskar von Stryk, Abdelhak~M. Zoubir, and Michael Muma.
\newblock Emergency response person localization and vital sign estimation using a semi-autonomous robot mounted sfcw radar.
\newblock \emph{IEEE Transactions on Biomedical Engineering}, 71\penalty0 (6):\penalty0 1756--1769, 2024.

\bibitem[Shokouhmand et~al.(2022)Shokouhmand, Eckstrom, Gholami, and Tavassolian]{2.13.3}
Arash Shokouhmand, Samuel Eckstrom, Behnood Gholami, and Negar Tavassolian.
\newblock Camera-augmented non-contact vital sign monitoring in real time.
\newblock \emph{IEEE Sensors Journal}, 22\penalty0 (12):\penalty0 11965--11978, 2022.

\bibitem[Tran et~al.(2019)Tran, Al-Jumaily, and Islam]{2.1.07}
Vinh~Phuc Tran, Adel~Ali Al-Jumaily, and Syed Mohammed~Shamsul Islam.
\newblock Doppler radar-based non-contact health monitoring for obstructive sleep apnea diagnosis: A comprehensive review.
\newblock \emph{Big Data and Cognitive Computing}, 3\penalty0 (1), 2019.
\newblock ISSN 2504-2289.

\bibitem[Wang et~al.(2021)Wang, Nguyen, Sridhar, and Gollakota]{2.1.03}
Anran Wang, Dan Nguyen, Arun~R. Sridhar, and Shyamnath Gollakota.
\newblock Using smart speakers to contactlessly monitor heart rhythms.
\newblock \emph{Communications Biology}, 4\penalty0 (1):\penalty0 319, March 2021.
\newblock ISSN 2399-3642.

\bibitem[Wang et~al.(2024)Wang, Wang, Zhang, Zhang, and Xu]{2.13.4}
Yingqi Wang, Zhongqin Wang, Jian~Andrew Zhang, Haimin Zhang, and Min Xu.
\newblock Vital sign monitoring in dynamic environment via mmwave radar and camera fusion.
\newblock \emph{IEEE Transactions on Mobile Computing}, 23\penalty0 (5):\penalty0 4163--4180, 2024.

\bibitem[Xiong et~al.(2020)Xiong, Peng, Gu, Li, Wang, and Zhang]{2.12.2}
Yuyong Xiong, Zhike Peng, Changzhan Gu, Songxu Li, Dong Wang, and Wenming Zhang.
\newblock Differential enhancement method for robust and accurate heart rate monitoring via microwave vital sign sensing.
\newblock \emph{IEEE Transactions on Instrumentation and Measurement}, 69\penalty0 (9):\penalty0 7108--7118, 2020.

\bibitem[Xu et~al.(2022)Xu, Zhang, Zhang, and Tao]{3.0.1}
Yufei Xu, Jing Zhang, Qiming Zhang, and Dacheng Tao.
\newblock Vitpose: simple vision transformer baselines for human pose estimation.
\newblock In \emph{Proceedings of the 36th International Conference on Neural Information Processing Systems}, NIPS '22, Red Hook, NY, USA, 2022. Curran Associates Inc.
\newblock ISBN 9781713871088.

\bibitem[Yang et~al.(2020)Yang, Bruce, Liu, Gholami, and Tavassolian]{2.13.1}
Chenxi Yang, Brendan Bruce, Xiaofan Liu, Behnood Gholami, and Negar Tavassolian.
\newblock A hybrid radar-camera respiratory monitoring system based on an impulse-radio ultrawideband radar.
\newblock In \emph{2020 42nd Annual International Conference of the IEEE Engineering in Medicine \& Biology Society (EMBC)}, pages 2646--2649, 2020.

\bibitem[Yang et~al.(2023)Yang, Huang, Zhou, Chen, Xu, Yuan, Zou, Lu, and Xie]{3666122.3666944}
Jianfei Yang, He~Huang, Yunjiao Zhou, Xinyan Chen, Yuecong Xu, Shenghai Yuan, Han Zou, Chris~Xiaoxuan Lu, and Lihua Xie.
\newblock Mm-fi: multi-modal non-intrusive 4d human dataset for versatile wireless sensing.
\newblock In \emph{Proceedings of the 37th International Conference on Neural Information Processing Systems}, NIPS '23, Red Hook, NY, USA, 2023. Curran Associates Inc.

\bibitem[Yuan et~al.(2026)Yuan, Kuo, Ren, Huo, Sun, and Zahid]{1.1.4}
Tongyang Yuan, Youyu Kuo, Aifeng Ren, Runqi Huo, Qing Sun, and Adnan Zahid.
\newblock State-of-the-art technologies and emerging trends in radar-based vital signs sensing: A review.
\newblock \emph{IEEE Sensors Journal}, 26\penalty0 (3):\penalty0 3446--3459, 2026.

\end{thebibliography}

\end{document}